\pdfoutput=1
\documentclass[11pt]{article}

\usepackage{acl}

\usepackage{times}
\usepackage{latexsym}
\usepackage{amssymb}
\usepackage{amsmath}

\usepackage[T1]{fontenc}
\usepackage[utf8]{inputenc}
\usepackage{tabularx}  % 引入tabularx包
\usepackage{multirow}  % 引入multirow包，以支持跨行文本
\usepackage{array}     % 引入array包，定义居中对齐的列类型
\usepackage{booktabs}
\usepackage{microtype}

\usepackage{inconsolata}

\usepackage{graphicx}

\title{\textit{LLM Critics Help Catch Bugs in Mathematics}: Towards a Better Mathematical Verifier with Natural Language Feedback}

\author{
 Bofei Gao\textsuperscript{1}, Zefan Cai\textsuperscript{1,4}, Runxin Xu\textsuperscript{1}, Peiyi Wang\textsuperscript{1}, Ce Zheng\textsuperscript{1}, Runji Lin\textsuperscript{2}, Keming Lu\textsuperscript{2} \\ \textbf{Dayiheng Liu\textsuperscript{2}, Chang Zhou\textsuperscript{2}, Wen Xiao\textsuperscript{3}, Junjie Hu\textsuperscript{4}, Tianyu Liu\textsuperscript{2} $\footnotemark[2]$, Baobao Chang\textsuperscript{1}} $\footnotemark[2]$ $\footnotemark[3]$ \\
 \textsuperscript{1} Peking University 
 \textsuperscript{2} Alibaba Group  \\
 \textsuperscript{3} Microsoft 
 \textsuperscript{4} University of Wisconsin - Madison \\
  \texttt{gaobofei@stu.pku.edu.cn, \{zefncai,wangpeiyi9979\}@gmail.com} \\
  \texttt{ tianyu0421@alibaba-inc.com, chbb@pku.edu.cn} \\
  \url{https://github.com/KbsdJames/MATH-Minos} \\
}

\begin{document}
\maketitle
\renewcommand{\thefootnote}{\fnsymbol{footnote}}
\begin{abstract}

In recent progress, mathematical verifiers have achieved success in mathematical reasoning tasks by validating the correctness of solutions generated by policy models. However, 
existing verifiers are trained with binary classification labels,
which are not informative enough for the model to accurately assess the solutions.
To mitigate the aforementioned insufficiency of binary labels, we introduce step-wise natural language feedback as rationale labels, that is, the correctness of each step and the detailed explanations.
In this paper, we propose \textbf{Math-Minos}, a natural language feedback-enhanced verifier by constructing automatically generated training data and a two-stage training paradigm for effective training and efficient inference.
Our experiments reveal that a small set of natural language feedback can significantly boost the performance of the verifier in both verification and reinforcement learning. We have released the code and data for further exploration.

\end{abstract}

\footnotetext[2]{Project Lead.}
\footnotetext[3]{Corresponding author.}
\renewcommand{\thefootnote}{\arabic{footnote}}

\section{Introduction}

\begin{figure}[htbp]
    \centering
    \includegraphics[width=1\linewidth]{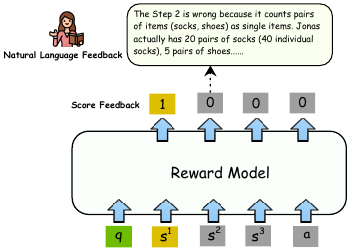}
    \caption{
    Illustration of \textbf{Score Feedback} and \textbf{Natural Language Feedback}.
    $q$ represents the mathematical questions
    $s_1, s_2, s_3$ represent the intermediate solutions. $a$ represents the final answer.
    $0$ and $1$ represent the score feedback.
    Our work aims to mitigate the insufficiency of score feedback and enhance verifiers' evaluation capabilities by introducing step-wise natural language feedback.
    }
    \label{fig:pre_study}
\end{figure}

Large Language Models (LLMs)~\cite{bai2023qwen,touvron2023llama,touvron2023llama2, jiang2023mistral, openai2024gpt4} have demonstrated strong capabilities in summarization~\cite{touvron2023llama2}, coding~\cite{rozière2024code}, tool using~\cite{song2023restgpt} and dialogue~\cite{ouyang2022training}.
% ~\citep{park2023generative, bai2023qwen, chen2023endtoend, zhang2023bayling, song2023restgpt}.
However, mathematical reasoning remains a challenge for LLMs~\cite{lightman2023lets, huang2024large}. To tackle this problem, recent research has focused on using verifiers to validate the correctness of response generated by models~\citep{wang2023selfconsistency, Zhu_2023, li-etal-2023-making,wang2024mathshepherd}.
An effective verifier can serve as 1) response reranker in the decoding~\cite{li-etal-2023-making, yu2024ovm, wang2024mathshepherd}; 2) reward model in further post-training~\cite{shao2024deepseekmath, wang2024mathshepherd}; 3) data purifier that filters erroneous responses in the SFT~\cite{rafailov2023direct}.

However, existing verifiers are all trained as binary classifiers by adding a classification head to an LLM. We argue that the score feedbacks as binary classification labels are not informative in training as they do not contain explanations for the underlying reasons for the errors, which causes inefficient training. 

In this work, we aim to enhance the verifier's evaluation ability for mathematical solutions by introducing step-level natural language feedback as rationale labels (i.e., the correctness of the current step and the further explanations).
We propose \textbf{MATH-Minos}, a natural language feedback enhanced verifier as shown in Section~\ref{sec:method}.
By employing supervised fine-tuning on only 30k natural language feedback before binary classification training, we can effectively enhance the model's evaluation capabilities.
In the first stage, we create high-quality step-level natural language feedback data as Section~\ref{subsec:data_construction}. To improve the quality of natural language evaluation, we introduce Label-aware Natural Language Feedback Curation. This approach simplifies the task by incorporating step-level binary classification labels, enhancing the evaluation generation process for GPT-4. 
Natural language feedback can provide in-depth reasons behind classification feedback, which enhances the training of the verifier.
In the second stage, we introduce a two-stage training for \textbf{MATH-Minos} as Section~\ref{subsec:model_training}: Firstly, we adopt the supervised fine-tuning on curated natural language feedback to effectively help improve the model's evaluation capabilities, followed by standard ORM \& PRM training on score feedbacks to achieve efficient inference with a single forward step.

The experiments in Section~\ref{section:experiments} demonstrate that infusing the model with evaluation capabilities via natural language feedback is more efficient and effective than traditional score feedback.
% Following \citet{wang2024mathshepherd}, we demonstrate the effectiveness of MATH-Minos in verification tasks. 
We show that only 30k training data with natural language feedback can significantly boost the performance of mathematical verifiers.
For Outcome Reward Model (ORM) setting, \textbf{MATH-Minos} improves the accuracy of MetaMath-Mistral~\citep{yu2024metamath} by 1.6\% (86.6\% $\rightarrow$ 88.2\%) on GSM8K and 0.7\% on MATH \% (37.6\% $\rightarrow$ 38.3\%) for ORM.
For Process Reward Model (PRM) setting, \textbf{MATH-Minos} improves the accuracy by 0.7\% (87.1\% $\rightarrow$ 87.8\%) on GSM8K and 0.8\% (37.8\% $\rightarrow$ 38.6\%) on MATH.
For reinforcement learning, \textbf{MATH-Minos} serves as a better reward model with an accuracy improvement by 1.8\% (81.8\% $\rightarrow$ 83.6\%) on GSM8K and 1.5\% (31.3\% $\rightarrow$ 32.8\%) on MATH.

In summary, our contributions are threefold:

1. We are the first study to conduct in-depth analyses of the reasons behind the evaluations generated by verifiers, revealing the shortcomings of the current verifier's training paradigm and inspiring future research.

2. We propose \textit{MATH-Minos} by proposing label-aware natural language feedback curation and a two-stage training paradigm, demonstrating that training verifiers with natural language feedback can complement the non-informative score feedback thus enhancing the model's evaluation ability.

3. We demonstrate the effectiveness of MATH-Minos across both ORM and PRM task settings. Extensive analysis demonstrates the superiority of the proposed method.

\section{Related Works}

\paragraph{Enhancing the mathematical reasoning ability of LLM}

Previous works focus on improving the mathematical reasoning ability of LLMs on three ways:
(1) Pre-training: LLMs~\cite{azerbayev2023llemma,openai2024gpt4,touvron2023llama,touvron2023llama2} are pre-trained on a large set of corpus related to mathematical questions with next-token prediction objective.
(2) Supervised fine-tuning:
Supervised fine-tuning can also improve the mathematical reasoning ability of LLMs by training LLMs with mathematical questions with detailed solutions ~\cite{yu2023metamath,luo2023wizardmath,wang2023making}.
(3) Inference: ~\cite{wei2022chain,fu2022complexity,zhang2023cumulative,bi2023program} design prompting strategies to improve the reasoning ability of LLMs.

\paragraph{Verifier for mathematical reasoning}

Previous mathematical verifiers can be mainly categorized into two categories:
Outcome Reward Model (ORM) gives an evaluation score to the whole solution; Process Reward Model (PRM) gives an evaluation score to each intermediate step of the solution.
Previous works~\cite{yu2023outcome,ying2024internlmmath,wang2024mathshepherd} use question-solution pair data with a score to train a ORM or a PRM, which is inefficient to help models understand the errors.
Therefore, in this work, we aim to train a verifiers with error types and detailed explanations about the errors.

\begin{figure*}[!]
    \centering
    \includegraphics[width=1\linewidth]{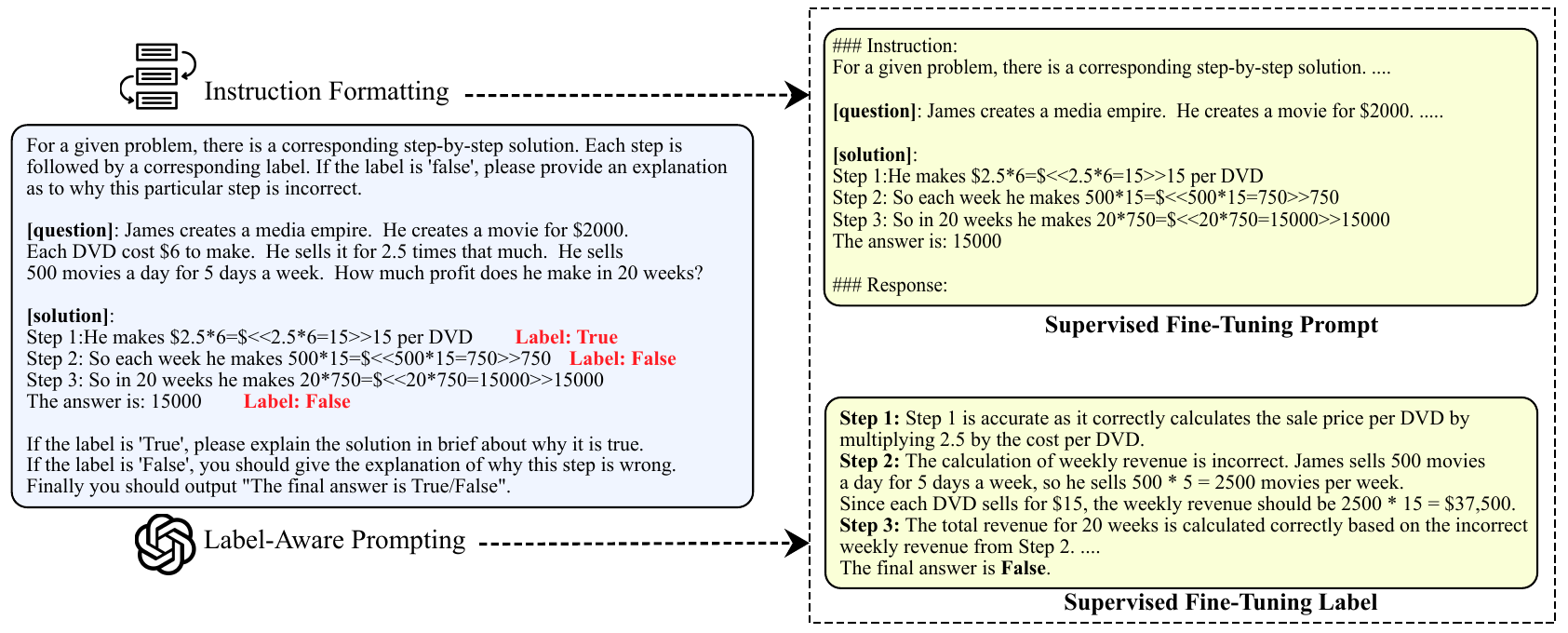}
    \caption{The illustration of the label-aware natural language feedback curation of MATH-Minos on GSM8K~\cite{cobbe2021training} dataset.
    We introduce step-level classification label to achieve step-level natural language feedback curation.
    }
    % \zc{Make the text larger.}
    % }
    \label{fig:data_creation}
\end{figure*}

\section{Methodology}
\label{sec:method}

In this section, we introduce the background of our proposed method (\S\ref{subsec:orm}), then delve into our proposed \textbf{MATH-Minos}, which contains label-aware natural language feedback curation (\S\ref{subsec:data_construction}) and the two-stage model training (\S\ref{subsec:model_training}).

\subsection{Background}
\paragraph{Outcome Reward Model}
\label{subsec:orm}
For a given problem $p$, the Outcome Reward Model (ORM) assigns a reward $r \in \mathbb{R}$ based on the whole completion $s$. The common approach for training an ORM involves implementing a binary sequence classification, which adds a classifier at the end of the LLM. The training loss is represented as follows:
\begin{align} \label{eq:orm}
    \mathcal{L}_{orm} = y_s \cdot log(\hat{y}_s) + (1-y_s) \cdot log (1 - \hat{y}_s),
\end{align}
where $y_s$ is the golden label of the solution and $\hat{y}_s$ is the sigmoid score of $s$ predicted by ORM.
For mathematical reasoning tasks, the quality of a sample can be directly determined by judging the correctness of the result. Therefore, the general approach to train a ORM involves using a generator to provide completions. Subsequently, rule-based matching is employed to determine the correctness of the current completion, and this outcome is used as the label for training. For the sake of simplicity and comparability, we directly modified the open-sourced dataset provided by \citet{wang2024mathshepherd} as the training set of ORM.

\paragraph{Process Reward Model}
\label{subsec:prm}
For a given question $q$, the Process Reward Model (PRM) assigns a reward $r \in \mathbb{R}$ to each step $s_i$ of the completion $s$. The training of PRM is through the task of token classification, the training loss can be represented as follows: 
\begin{align} \label{eq:prm}
\scalebox{0.9}{
$
\mathcal{L}_{prm} = \sum\limits_{i=1}^{K}y_{s_i} \cdot log(\hat{y}_{s_i}) + (1-y_{s_i}) \cdot log (1 - \hat{y}_{s_i}),
$
}
\end{align}
where $K$ is the number of reasoning steps of the completion, $y_{s_i}$ is the golden label of the solution and $\hat{y}_{s_i}$ is the sigmoid score of $s_i$ predicted by PRM. Compare to ORM, PRM can provide fine-grained supervision which is more detailed and reliable.

\label{subsec:model_training}
\begin{figure*}[!]
    \centering
    \includegraphics[width=1\linewidth]{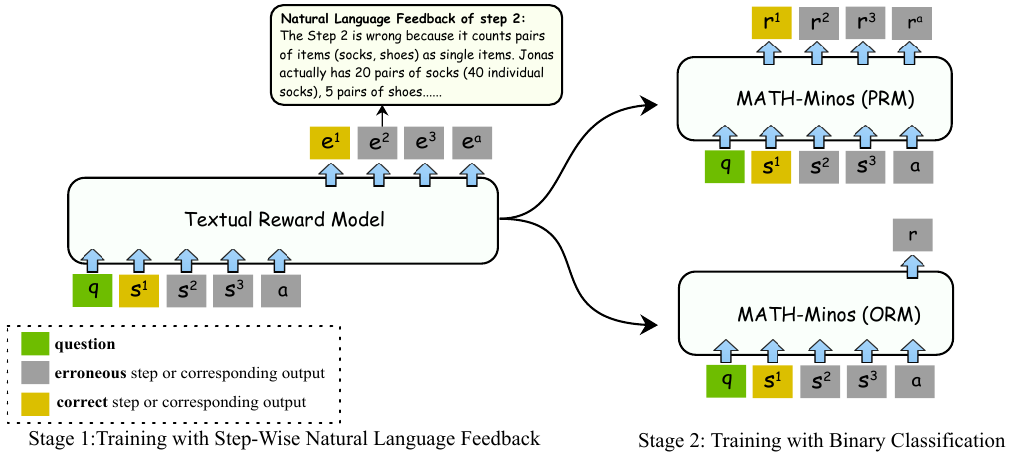}
    \caption{The overview of the two-stage training process of MATH-Minos.
    In Stage 1, training reward model (RM) with natural language feedbacks helps RM learn to evaluate effectively and efficiently.
    In Stage 2, training RM as binary classification helps RM inference efficiently by outputing a reward with one single forward pass.
    }
    \label{fig:two_stage}
\end{figure*}

\subsection{Label-aware Natural Language Feedback Curation}
\label{subsec:data_construction}

\begin{table}
\begin{tabularx}{0.95\linewidth}{lXX} % 使用tabularx环境，设置表格宽度为\linewidth，即占满单栏
\toprule
\textbf{Task} & \textbf{GSM8K} & \textbf{MATH} \\
\midrule
Outcome Evaluation     &  \quad95.1              &    \hspace{5pt} 62.0           \\
Process Evaluation        & \quad85.1               &   \hspace{5pt} 59.7            \\
\bottomrule
\end{tabularx}

\caption{The step-level Acc. score of prompting GPT-4 to generate natural language feedback.}
\label{tab:data_generation}

\end{table}
In this section, we introduce the label-aware natural language feedback curation of MATH-Minos as shown in Figure \ref{fig:data_creation}. Since the natural language feedback can be understood by both humans and large models, it is suitable to stimulate the evaluation capabilities of LLMs. Unlike the binary label, natural language feedback provides detailed explanations for right or wrong completions, which also brings complexity to data collection. The best way for generating the natural language feedback data is through manual annotation. Considering the costs associated with human annotation, we obtain natural language feedback by leveraging the capabilities of the GPT-4-turbo~\citep{openai2024gpt4}. 

To verify the quality of data, we sample data from the Math-Shepherd~\cite{wang2024mathshepherd} and PRM800K~\cite{lightman2023lets} to create an evaluation dataset including 500 question-solution samples with step-level and outcome-level binary classification label for GSM8K~\cite{cobbe2021training} and MATH~\cite{hendrycks2021measuring}. We then check the accuracy of outcome evaluation and step-level evaluation, with results presented in Table \ref{tab:data_generation}. Experimental results show that this prompting manner doesn't yield high-quality data with only 85.1 step-level accuracy for GSM8K and 59.7 step-level accuracy for MATH. This also indicates that one of the factors limiting the performance of the reward model is the base model's evaluation capability.

To facilitate GPT-4 in generating higher-quality data, we propose a label-aware prompting method, which simplifies the evaluation task by introducing the binary classification label within the prompt. As illustrated in Figure~\ref{fig:data_creation}, GPT-4's task shifts from determining correctness and generating explanations to generating explanations based on the given label. 
After obtaining automatically labeled data, we hire a group of graduate students to review the data, further ensuring its quality.
% Extensive analysis in Section~\ref{sec:anaysis} have also validate the effectiveness of our approach.

\subsection{Two-Stage Training of MATH-Minos}
\label{subsec:model_training}

Based on the aforementioned data generated in Section ~\ref{subsec:data_construction}, we introduce a novel two-stage training paradigm including
(1) Stage 1: Training with Step-Wise Natural Language Feedback and
(2) Stage 2: Training with Binary Classification
to synergistically combine the strengths of evaluation generator and discriminator, which is shown in figure \ref{fig:two_stage}.
This training paradigm enjoys two potential benefits:
Firstly, natural language feedback contains rich information, especially for complex reasoning tasks such as mathematics. Therefore training with natural language feedback can significantly improve the models' evaluation ability with just a small set of data.
Secondly, the inference of binary classification discriminator is more efficient compared with natural language feedback generation. This approach not only allows model to generate a score but also enables the model to produce evaluation results with just a single forward pass, thereby enhancing the efficiency.

\paragraph{Reward Modeling with Natural Language Feedback}
In the first stage, we employ supervised fine-tuning to enhance the evaluation capabilities of the model. We utilize the \textit{Supervised Fine-Tuning Prompt} shown in Figure \ref{fig:data_creation} as the input $x_{q, s}$ for the model, with the \textit{Supervised Fine-Tuning Label} generated by GPT-4 serving as the model's output $y$. The training loss for a sample can be defined as follows: 
\begin{align} \label{eq:train}
    L_{textrm} = \sum_{t=1}^{M}log P(y_t | y_{< t}, x_{q, s}),
\end{align}
where $M$ is the total length of $y$ and $y_{< t}$ is the previous tokens.

% Empirical evidence demonstrates that this approach, in comparison to binary classification, effectively augments the evaluation capabilities of LLMs with only a minimal dataset.

\paragraph{Reward Modeling with Binary Classification}
% \zc{Add a classificaiton loss here}
After the first stage, the evaluation capability of the model is improved. However, natural language feedback cannot provide a score and thus can't be used as a reward for further optimizations like PPO~\cite{schulman2017proximal}. Additionally, when the model generates feedback, it produces a complete evaluation with rationales, making it significantly less efficient than using a classification-based verifier. Therefore, we further train the verifier with binary classification labels as \autoref{eq:orm} and \autoref{eq:prm}.

Benefiting the proposed two-stage training, we can enhance the verifier's evaluation ability with natural language feedbacks and efficiently apply the verifier to PRM or ORM with one single forward pass.
% conveniently adapt the textual reward model to PRM or ORM.
% Experiment show that the reward model converges faster for whether PRM or ORM. 

\section{Experiment}
\label{section:experiments}

\subsection{Experiment Setup}

\newcolumntype{Y}{>{\centering\arraybackslash}X}
\begin{table*}[ht]
\resizebox{0.97\linewidth}{!}{
\begin{minipage}{1.06\linewidth}
\begin{tabularx}{\textwidth}{l l Y Y}  % 只将最后两列定义为X类型
\toprule
\textbf{Models}                       & \textbf{Verifier}              & \textbf{GSM8K} & \textbf{MATH500} \\ \hline
\multirow{9}{*}{Mistral-7B: MetaMATH} & Self-Consistency~\cite{li-etal-2023-making}              & 84.1           & 34.6             \\
                                      & ORM~\cite{wang2024mathshepherd}                           & 86.2           & 35.9             \\ 
                                      & PRM~\cite{wang2024mathshepherd}                           &  87.1              &  36.7                \\ 
                                      & Self-Consistency + ORM~\cite{ wang2024mathshepherd}       & 86.6           & 37.6             \\ 
                                      & Self-Consistency + PRM~\cite{wang2024mathshepherd}        & 86.8           & 37.8             \\ \cline{2-4} 
                                      & MATH-Minos (ORM) \textsuperscript{$\dagger$}          & 87.3               & 37.4                 \\
                                      & MATH-Minos (PRM) \textsuperscript{$\dagger$}           & 87.6             &  37.8                 \\
                                      & Self-Consistency + MATH-Minos (ORM) \textsuperscript{$\dagger$}       &  \textbf{88.2}    &  38.3   \\
                                      & Self-Consistency + MATH-Minos (PRM) \textsuperscript{$\dagger$}       &  87.8     &   \textbf{38.6}    \\ 
                                      \bottomrule
\end{tabularx}
\end{minipage}
}
\caption{Main results of MATH-Minos in verification. The verification is based on 256 outputs. $\dagger$ denotes the method is proposed in this paper.
Our \textbf{MATH-Minos} significantly outperforms baselines in both ORM and PRM settings.
}
\label{tab:main_result}
\end{table*}

\paragraph{Dataset} We conduct our experiment on two widely used mathematical datasets GSM8K~\cite{cobbe2021training} and MATH~\cite{hendrycks2021measuring}. GSM8K\cite{cobbe2021training} comprises a variety of word problems that are typically found in grade school mathematics curricula, which contains 7473 samples in the training set and 1319 samples in the test set. MATH ~\cite{hendrycks2021measuring}is a diverse collection of mathematical problems that cover a broad range of topics and skill levels, from elementary to advanced mathematics, which contains 7500 samples in the training set and 5000 samples in the test set. In the setting of verification, we sample the test set of MATH to 500 samples which is identical to \citet{lightman2023lets}.

\paragraph{Verification} Following \citet{lightman2023lets} and \citet{wang2024mathshepherd}, we adopt the best-of-N selection to evaluate the capability of our verifier. Specifically, given a question $q$ and a generator, we let the generator sample $N$ times for the question $q$. Then, the verifier is used to evaluate the quality of each completion. The final answer $a$ is determined as the one with the highest reward according to the verifier's output $ RM(q, a_i)$, formally expressed as follows:
\begin{equation}
    a_{\text{rm}} = \mathcal{F} (\arg\max_{s_i} RM(q, s_i)),
\end{equation}
where $s_i$ is the i-th solution generated by generator and $\mathcal{F}(\cdot)$ denotes extracting the final answer from the solution.

Following \citet{li-etal-2023-making} and \citet{wang2024mathshepherd}, we also explore the ensemble of self-consistency (majority voting) and the verifier. Specifically, we classify the results output by the model into different groups and calculate the cumulative reward for each group, which can be calculated as follows:
\begin{equation}
\text{\scalebox{0.9}{$
    a_{\text{sc+rm}} = \arg\max\limits_{a} \sum_{i=1}^{N} \mathbb{I}(\mathcal{F}(s_i) = a) \cdot RM(q, s_i),
$}}
\end{equation}
where $N$ is the number of solutions, $s_i$ is the solution generated by generator and $\mathcal{F}(\cdot)$ denotes extracting the final answer from the solution.

\paragraph{Reward Model} Following~\citet{wang2024mathshepherd}, to further examine the performance of the verifier, we evaluate it as the reward model for PPO \citep{schulman2017proximal} during the post-training phase. An effective reward model can provide reliable feedback signals, thereby enhancing the final performance of PPO.

\paragraph{Experimental Setting} For the training of the verifiers, to ensure comparability and convenience, we utilize the open-source MATH-shepherd dataset~\cite{wang2024mathshepherd} for both baseline reward models and MATH-Minos. We curate a total of 30K samples of natural language feedback using the data from phase-one of PRM800K~\cite{lightman2023lets} and the subset of MATH-Shepherd. Our main experiment conducts the verification on the test set of GSM8K and MATH and reinforcement learning (PPO) on the training set of GSM8K and MATH.
We follow \citet{wang2024mathshepherd} with their detailed experimental setup, as shown in Appendix~\ref{sec:detailed_experimental_setup}.

\begin{table}

\small
\begin{tabularx}{\linewidth}{lXX} % 使用tabularx环境，设置表格宽度为\linewidth，即占满单栏
\toprule
Model & \textbf{GSM8K} & \textbf{MATH500} \\
\midrule
Mistral-7B: MetaMATH       & 77.9                   &  28.6                             \\
+ ORM-PPO    & 81.8             &  31.3            \\
+ MATH-Minos(ORM)-PPO     & \textbf{83.6}             &   \textbf{32.8}            \\
\bottomrule
\end{tabularx}

\caption{
Experimental result of the MATH-Minos in PPO post-training setting. Our MATH-Minos significantly outperforms the vanilla ORM on PPO training.
}
\label{tab:ppo}

\end{table}
\subsection{Main Result}
We present the performance of various methods in Table~\ref{tab:main_result} and Table~\ref{tab:ppo}. 

For the task of verification, MATH-Minos (ORM) achieves an improvement of 1.1\% in accuracy on the GSM8K and 0.7\% in on the MATH compared to traditional ORM. For PRM, MATH-Minos (PRM) achieves an accuracy improvement of 0.7\% on GSM8K and 0.8\% on the MATH. Ensembling with self-consistency and MATH-Minos, the MetaMATH-Mistral generator achieves optimal accuracy of 88.2\% on GSM8K and 38.6\% on MATH500. Beyond the improvement of the performance, we find that MATH-Minos has a more pronounced effect in the setting of ORM. We believe this phenomenon could be attributed to the sparser supervision in ORM compared to PRM, implying that information-rich textual explanations can offer more substantial benefits to ORM.

For the task of reward model on reinforcement learning, MATH-Minos (ORM) achieves an improvement of 1.8\% in accuracy on the GSM8K and 1.5\% in on the MATH compared to traditional ORM, demonstrating the robust effectiveness of our method across various task settings.
%证明了我们的方法在多种任务设置上均有效。
%  1) Compared with GSM8K, MATH-Minos achieves greator improvement on MATH. 
% This showcases that training with natural language feedback can effectively impart knowledge to the model, significantly boosting their capability to evaluate more complex reasoning tasks. Especially for MATH that necessitate multi-steps and logical reasoning, the demand for explanations is heightened.

% 2) 

\section{Analysis}
\label{sec:anaysis}
\subsection{Error Distribution of Math Solvers}
\label{sec:err_distributions}
To further illustrate the shortcomings of training the verifier solely relying on binary classification, we conduct an in-depth investigation into the errors produced by the generator at the step level. Specifically, we take the natural language feedback generated by GPT-4 as a reference and heuristically categorize the causes of errors in responses into five distinct types:
\textit{\textbf{Unrelated}}
\textit{\textbf{Unrelated}}: This indicates that the step is irrelevant and does not contribute towards deducing the final answer.
\textit{\textbf{Accumulation}}: This denotes that the step is incorrect due to a mistake in the preceding step, leading to subsequent errors.
\textit{\textbf{Calculation}}: This categorization is reserved for errors arising from incorrect calculations, which is one of the most common errors in mathematical reasoning.
\textit{\textbf{Logic}}: This applies to steps that are logically flawed in the context of solving the given problem.
\textit{\textbf{Other}}: This category encompasses steps that are erroneous for reasons not covered by the aforementioned categories.

\begin{figure}[!]
    \centering
    \includegraphics[width=0.9\linewidth]{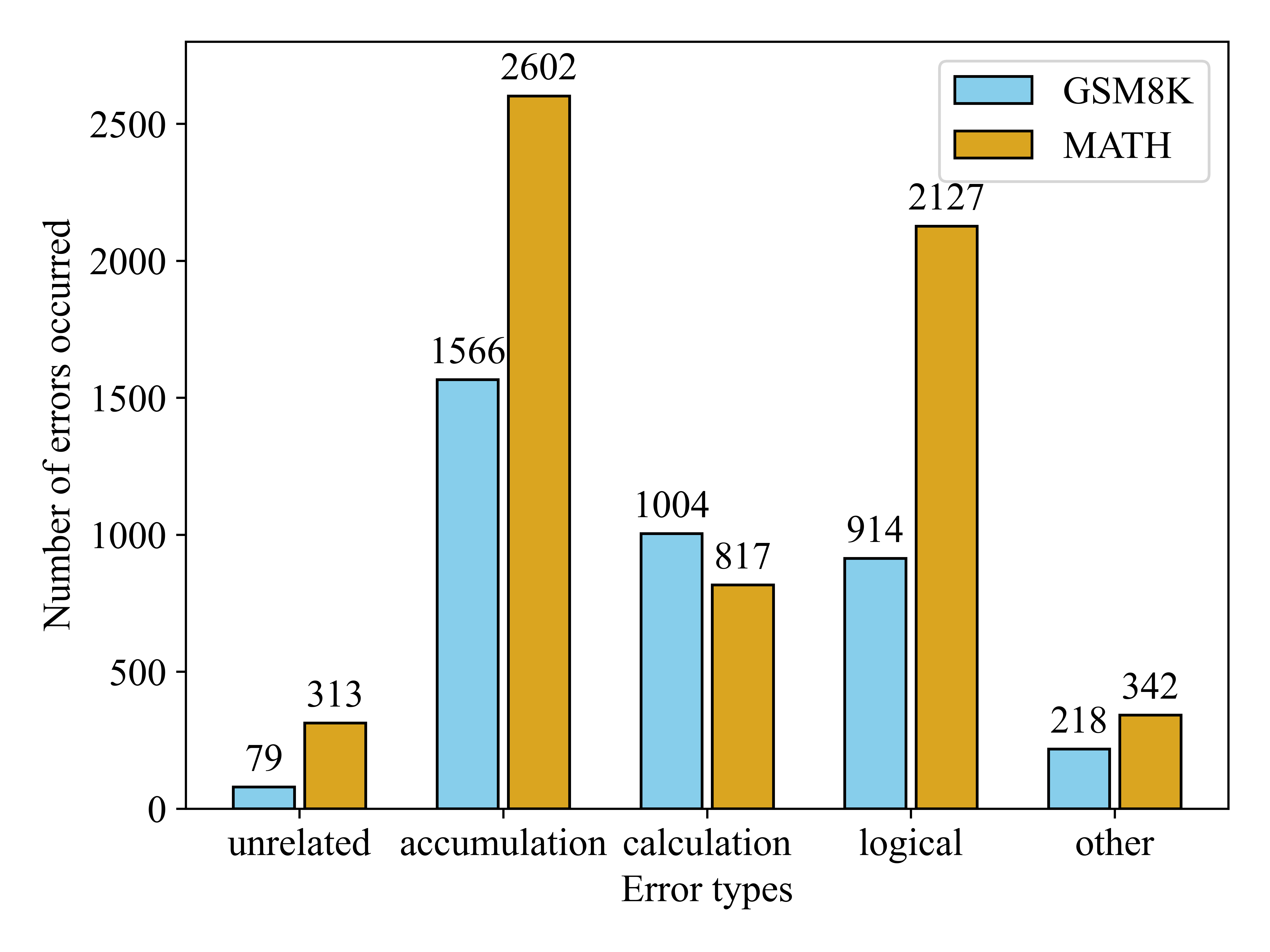}
    \caption{Statistics on the types of reasoning errors of MetaMath-Mistral on the GSM8K and MATH.}
    \label{fig:error_analysis}
\end{figure}

We use the same way as the label-aware prompting introduced in Section~\ref{fig:data_creation} to automatically analyze the cause of errors. We obtain the step-level labels from the subset of MATH-Shepherd~\citep{wang2024mathshepherd}, which contains 2500 samples for both GSM8K and MATH. Given the question, solution and the natural language feedback, we employ GPT-4 for the classification of error causes. The experimental result is shown in Figure~\ref{fig:error_analysis}.

From our statistical analysis, it is evident that the model produces errors across all types. For the MATH dataset, given its higher difficulty level and the necessity for more steps, a greater total number of errors occur within the same number of samples of GSM8K.
Furthermore, we discover that the most common cause of errors in both datasets is accumulation, which is consistent with our intuition. In multi-step reasoning, a mistake in one step is likely to directly cause errors in all subsequent steps. Furthermore, we observe distinct patterns of errors in the GSM8K and MATH datasets. For the GSM8K dataset, the occurrences of calculation errors and logical errors were approximately the same. Instead, in the MATH dataset, logical errors significantly outnumber calculation errors. This also indirectly demonstrates that models are vulnerable to more complex reasoning tasks.

These findings further illustrate that using binary labels to supervise the learning of reasoning evaluation tasks is insufficient and therefore highlighting the proposal for using natural language feedback to supplement the training of vanilla ORM or PRM.

\subsection{Meta-Evaluation and Convergence Curves}
\begin{figure}[!ht]
    \centering
    % 第一张图片
    \includegraphics[width=\linewidth]{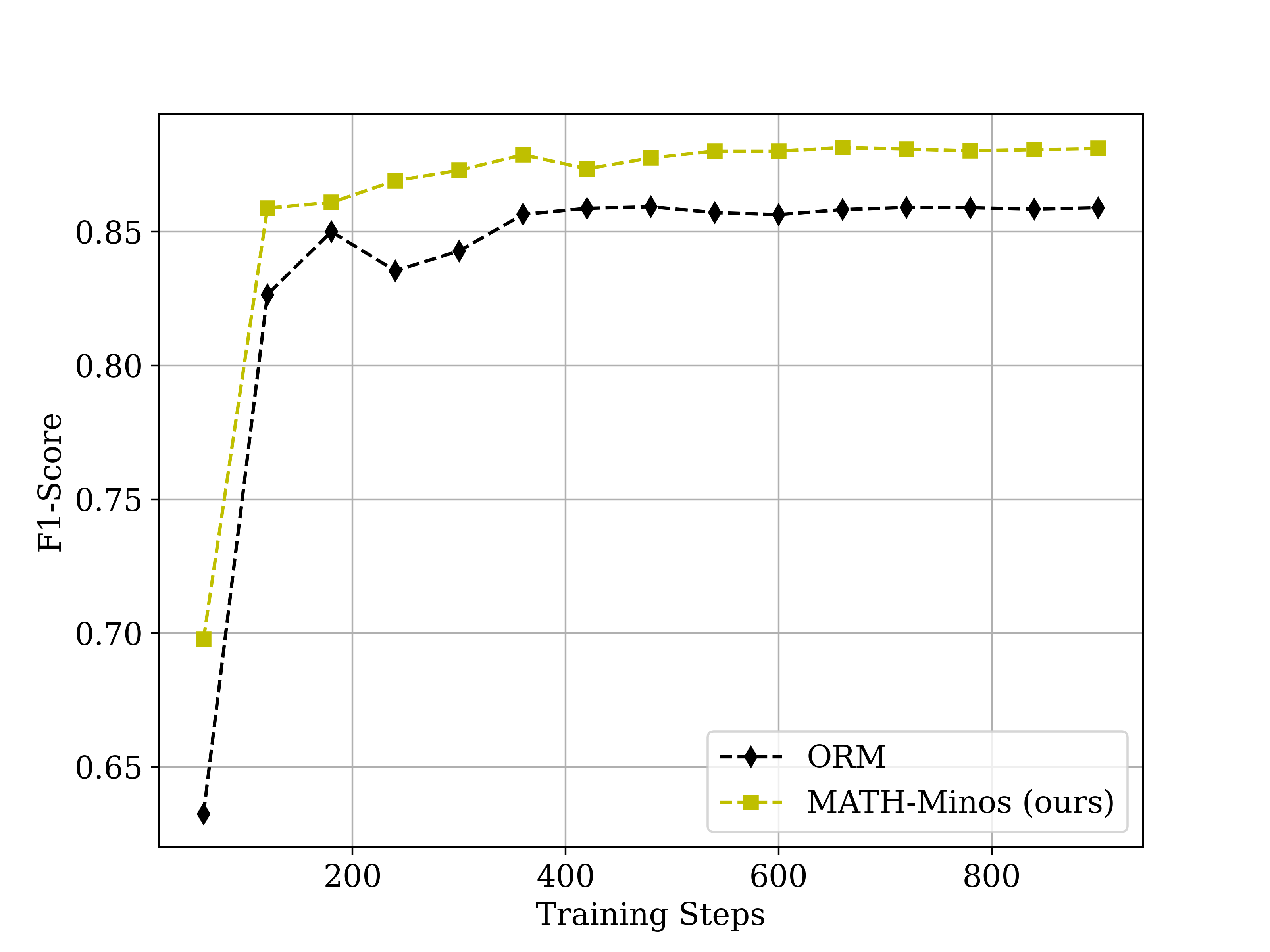}
    \caption{
    The convergence curve of vanilla ORM and our \textbf{Math-Minos} with natural language feedbacks on meta-evaluation set.
    }
    \label{fig:convergence}

    % \vspace{1em} % 为两张图片添加一些垂直间距

    % 第二张图片
    \includegraphics[width=0.9\linewidth]{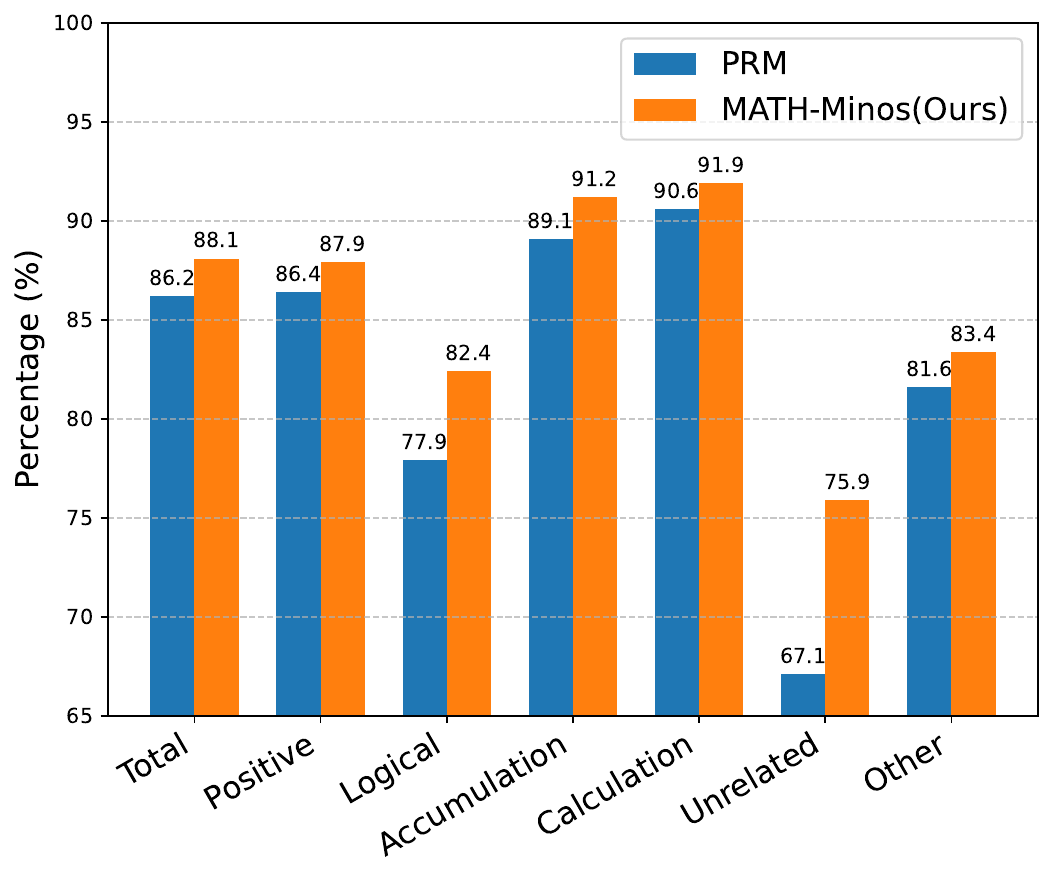} % 替换成你第二张图片的文件名
    \caption{The detailed improvement of the step-level model prediction accuracy between vanilla PRM and \textbf{Math-Minos}.}
    \label{fig:detailed_improve}
\end{figure}

To measure the verifier's capabilities in a more convenient and direct manner instead of verification, an intuitive approach is to assess whether the verifier can accurately determine the correctness of the final answer. Without the influence of the generator, this method purely relies on the capability of the verifier. We construct a meta-evaluation set based on whether the final answer provided in the generator's output is correct, serving as the ground truth label. By sampling several responses from the generator on the test set and deriving labels through rules, we create a meta-evaluation set for GSM8K containing 20,000 samples. We conduct tests on the meta-evaluation set using the checkpoints of each epoch after completing the model training and verification. The results of the meta-evaluation are presented in Figure \ref{fig:convergence}.
Additionally, we construct the step-level meta-evaluation dataset using the same data in Section \ref{sec:err_distributions} to provide insights into where our method's gains are.

We observe that MATH-Minos consistently outperforms vanilla ORM in meta-evaluation. Additionally, MATH-Minos exhibit a faster convergence rate, surpassing the baseline at only approximately 120 steps. Given that we trained for only 1 epoch, this means that in the actual secondary phase of binary classification, only about 60K data are required to exceed the baseline. Hence, the experiment demonstrates that natural language feedback can significantly reduce the amount of data needed to train a verifier.

Furthermore, in the step-level meta-evaluation as Figure \ref{fig:detailed_improve}, we can observe that our method significantly aids in identifying all types of errors, with the most notable improvements seen in logical errors and unrelated contents. This aligns well with the strengths of natural language feedback, which provides a more detailed analysis of the underlying reasons compared to score-based feedback. Such insights help enhance the logical reasoning capabilities of the reward model and effectively identify redundant information.

\subsection{Influence of the Feedback Quality}
\label{sec:analysis_quality}
\begin{table}

\small
\begin{tabularx}{\linewidth}{lXX} % 使用tabularx环境，设置表格宽度为\linewidth，即占满单栏
\toprule
GSM8K & \textbf{Meta-Eval} & \textbf{Verification} \\
\midrule
ORM        & 85.9                   &  86.2                             \\
+ curation w/o label    & 86.2             &  85.8            \\
+ curation w/ label     & 88.1             &   88.2            \\
\bottomrule
\end{tabularx}

\caption{
Experimental result of the ORM, ORM with vanilla natural language feedback curation and our Math--Minos (ORM with label-aware natural language feedback curation).
Label-aware natural language feedback curation significantly enhances ORM's evaluation ability.
}
\label{tab:data_quality}

\end{table}

\begin{table}

\small
\begin{tabularx}{\linewidth}{lXX} % 使用tabularx环境，设置表格宽度为\linewidth，即占满单栏
\toprule
GSM8K & \textbf{GSM8K} & \textbf{MATH} \\
\midrule
Vanilla ORM     & 86.2                  &  35.9 \\
MATH-Minos (GPT-4)       & 	87.3            &  37.4      \\
MATH-Minos (Qwen2-72b)    & 87.5             &  36.8            \\
MATH-Minos (Mistral-7B)     & 85.4        &   34.5    \\
\bottomrule
\end{tabularx}
\caption{
Experimental result of the ORM, Math-Minos with label-aware natural language feedback curation from different models. "Qwen2-72b" denotes we use Qwen2-72b-instruct as the natural language feedback generator. "Mistral-7B" denotes we use Mistral-7b-Instruct-v0.3 as the natural language feedback generator.
}
\label{tab:different_basemodel}

\end{table}
In this section, we conduct an extensive analysis of the impact on the data quality of natural language feedback.
One baseline is direct prompting of GPT-4-turbo. We use the 30K direct GPT-4 evaluation which is the same number of MATH-Minos to compare. We use both meta-eval and verification to measure the capability of the verifier. 
Additionally, to enhance the scalability and accessibility of our method, we explore the use of different base models to generate natural language feedback. Specifically, we utilize the open-source model Qwen2-72b-Instruct as a substitute for GPT-4-turbo. Furthermore, we take a more radical approach by directly employing the identical base model, Mistral-7B-Instruct-v0.3, which shares the same base model of our policy model MetaMATH: Mistral-7B. This allows us to comprehensively investigate the impact of the natural language feedback generated by these models on the verifier's performance.The experimental result is shown in Table~\ref{tab:data_quality} and Table~\ref{tab:different_basemodel}.

The experimental results indicate that directly prompting GPT does not significantly enhance the performance of the verifier. This is possibly due to the poor quality of the data shown in Table \ref{tab:data_generation}. From the experimental results in Table~\ref{tab:different_basemodel}, we observed that Qwen2-72b-Instruct can produce high-quality natural language feedback, leading to a similar enhancement of GPT-4-turbo in the verifier's performance, thereby demonstrating the scalability and accessibility of our approach. However, we found that Mistral-7B-Instruct-v0.3, which shares the same base model as our verifier, fails to improve the model's performance effectively. We believe this may be due to the model's smaller size, which hinders its ability to generate accurate natural language feedback, ultimately resulting in a decline in the verifier's performance.

\subsection{Ablation Study}
\begin{table}

\small
\begin{tabularx}{\linewidth}{lXX} % 使用tabularx环境，设置表格宽度为\linewidth，即占满单栏
\toprule
GSM8K & \textbf{Meta-Eval} & \textbf{Verification} \\
\midrule
ORM  w/o stage 1        & 85.7                   &  86.2                     \\
RM w/o stage 2    & 82.8             &  84.7            \\
MATH-Minos     & 88.0             &   88.2            \\
\bottomrule
\end{tabularx}

\caption{Ablation study of the two-stage training paradigm.
RM w/o stage 1 denotes that we only train the verifier with the score feedback.
RM w/o stage 2 denotes that we only train the verifier the natural language feedback generated.}
\label{tab:ablation_study}

\end{table}
Table~\ref{tab:ablation_study} presents the results of our ablation study, wherein we delve into the effect of each stage. 

Removing stage 1 essentially reverts our method to a vanilla ORM, where the model is unaware of the reasons behind what makes an answer correct or incorrect. Hence, the performance noticeably declines compared to MATH-Minos. 

When eliminating stage 2, binary classification, an intuitive approach is to directly utilize the natural language feedback generated by the text reward model for the generator's verification. Given that the model outputs a binary discrete value ('True' or 'False'), we cannot employ a best-of-N verification but instead use it as a filter. Specifically, we apply self-consistency in filtering out cases where the model output is 'True'. Unfortunately, we observe that relying solely on natural language feedback from the textual reward model leads to a significant decline in performance. The probable reasons may include: 1) Upon closer inspection, we notice inconsistencies in the model's feedback. This is characterized by samples where a step is recognized as incorrect yet the overall outcome is deemed correct, and vice versa. Such inconsistencies are even found in the strongest models such as GPT-4, despite their ability to provide accurate explanations. 2) The performance of evaluation might be constrained by the model's scale. Influenced by computational resources, we do not further explore larger models. Evaluation generation tasks could be more challenging for models of smaller scale. 3) The binary discrete output of the model is relatively coarse-grained. For instance, two examples judged as correct cannot be compared with each other.

In summary, in this section, our experiments demonstrate that both stage one and stage two are essential, where natural language feedback and binary classification play complementary roles.

\section{Conclusion}
%\wx{better to change a word, like 'investigate the effectiveness of'} 
We investigate the effectiveness of the current training paradigm of verifiers, demonstrating that score feedback from binary classification labels is suboptimal for teaching LLMs to accurately evaluate mathematical solutions. By introducing natural language feedback with a two-stage training paradigm, we significantly enhance the verifier's evaluation ability. The experimental results show that models trained on a small dataset with natural language feedback (30k instances) significantly outperform the baselines that rely solely on classification labels.
This highlights the critical role of rich and informative labels in training data in crafting more nuanced and effective training strategies for the development of LLMs that are capable of complex reasoning tasks. 
Finally, the findings of this work pave the way for the potential integration of natural language feedback with classification verifiers. 

%\wx{better to add one or two sentences about future work - how these findings and conclusions can motivate or lead the following research?}

\section{Limitations}
Following the scaling laws, the evaluation ability of a model, especially in terms of generating natural language evaluations, may vary across different sizes. However, due to computational resource limitations, experiments were conducted solely on a model with 7 billion parameters, thereby being unable to explore the impact of the model's scaling on the evaluation ability. We leave it to our future work. 
We do not verify the Math-Minos (PRM) in the reinforcement learning setting because of the lack of open-source implementation of step-by-step PPO. We also leave it to our future work.

\bibliography{custom}

\clearpage
\appendix
\section{Performance on False Positive Samples}
\label{sec:analysis_base_model}
\begin{table}

\small
\begin{tabularx}{0.99\linewidth}{lXX} % 使用tabularx环境，设置表格宽度为\linewidth，即占满单栏
\toprule
\quad & \textbf{Recall} & \textbf{Avg. Reward} \\
\midrule
ORM        & \hspace{3pt}0.74                   &  \quad\quad0.234             \\
MATH-Minos (ORM)    & \hspace{3pt}0.92            &  \quad\quad0.105            \\
\bottomrule
\end{tabularx}

\caption{The recall and average reward of the false positive examples (i.e., the final answer of the solution is true while the intermediate steps are false) of ORM and \textbf{MATH-Minos}.
\textbf{MATH-Minos} significantly improves the evaluation towards false positive examples.
}
\label{tab:false_positive}

\end{table}
To further investigate the efficacy of Math-Minos, we analyze the performance on false positive samples within the training dataset. False positive samples refer to those instances that have a correct final outcome but contain errors in the intermediate steps. Ideally, a robust verifier should assign lower rewards to these samples. We extract such examples using the step-level labels from the training set of the verifier, amounting to a total of 600 samples, which includes data from both GSM8K and MATH datasets. We test the performance of both ORM and Math-Minos, with the experimental results presented in Table~\ref{tab:false_positive}.

According to our experimental findings, it turns out that vanilla ORM can correctly discriminate a majority of the false-positive samples from the training set but with an accuracy significantly lower than MATH-Minos. It achieves a recall of 74\% with an average reward of 0.234. While MATH-Minos reach a recall rate of 92\% with an average reward of 0.105. This performance is significantly better than that of ORM not trained on the natural language feedback.

Delving into the data, we discover that in the context of false positives, a substantial portion of the natural language feedback generated by GPT-4 are contradicted to the “True” labels we assigned. We believe that these data endows MATH-Minos with a stronger capability to discern false-positives, thereby enhancing the model's performance.

\section{Influence of the Data Amount}
\begin{figure}[!]
    \centering
    \includegraphics[width=1\linewidth]{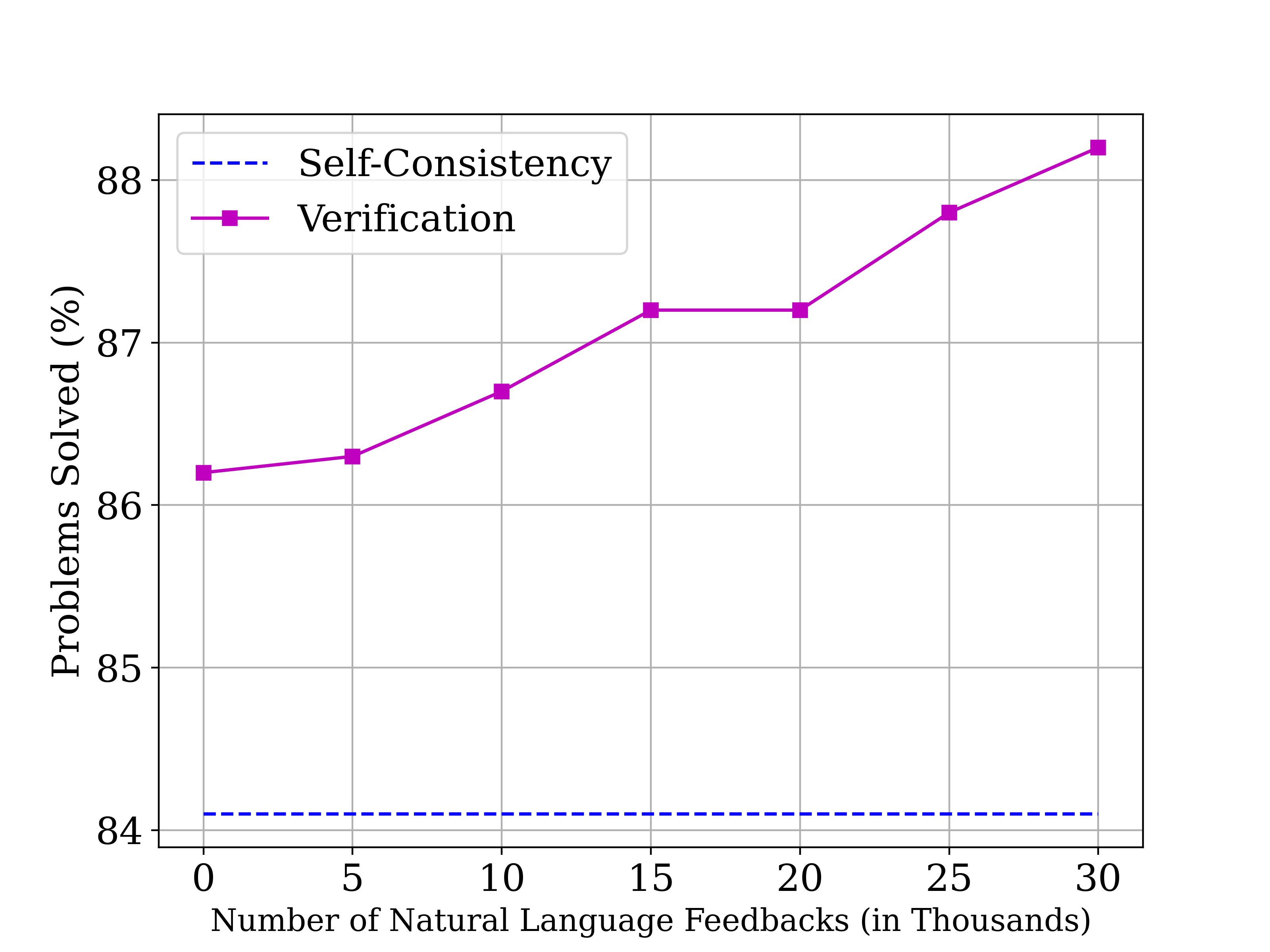}
    \caption{
    The impact of different amount of natural language feedback on the performance of the verifier in GSM8K.
    This shows the scalable potential of our \textbf{MATH-Minos}.
    }
    \label{fig:data_number}
\end{figure}

Figure~\ref{fig:data_number} illustrates how different amounts of natural language feedback affect the performance of MATH-Minos during the first stage. We use SFT on the model in the first stage using different scales of natural language feedback. In the second stage, we adopt the setup of ORM setting and use the verifier to select the best-of-N of GSM8K test set. We observe a positive correlation between the model's performance and the quantity of natural language feedback provided in stage one, which implicitly evidences the benefit of natural language feedback for the model.

\section{Comparison with COT distillation}
\begin{table}

\small
\begin{tabularx}{\linewidth}{lXX} % 使用tabularx环境，设置表格宽度为\linewidth，即占满单栏
\toprule
Model & \textbf{GSM8K} & \textbf{MATH500} \\
\midrule
ORM       & 86.2                   &  35.9         \\
Math-Minos(ORM)    & \textbf{87.3}             & \textbf{37.4}       \\
ORM w/ GPT-4-COT-Distillation    &  85.8    & 36.2  \\
\bottomrule
\end{tabularx}

\caption{
Experimental result of different ways incorporating GPT-4 chain-of-thought data. "ORM w/ GPT4-COT-Distillation" denotes we use two-stage training of ORM with direct COT distillation generated by GPT-4.
}
\label{tab:cot-baseline}

\end{table}
To further validate the efficacy of our method, we compared it against a more generalized baseline approach under conditions where the same number of GPT-4 tokens is utilized. Specifically, we employed enhanced queries from Meta-MATH and consequently distilled the problem-solving processes generated by GPT-4 into the ORM training. We aimed to determine whether training the model with COT data distilled using the same number of GPT-4 tokens could yield comparable results with MATH-Minos.

The experimental results are as follows: Our findings indicate that using an equivalent number of GPT-4 tokens 
 on COT-distillation resulted in negligible improvements to the verifier. This lack of enhancement can be attributed to the fact that its base model, Mistral-7B:MetaMATH, has already been trained on a substantial amount of high-quality COT data. Consequently, this step contributed almost no additional informational gain to the model. These results further underscore the effectiveness and necessity of our Label-Aware Natural Language Feedback Curation.

\section{Detailed Experimental Setup}
\label{sec:detailed_experimental_setup}
We use the MetaMATH-Mistral as the generator for the questions in the test set. In order to ensure the model has the ability to solve mathematical problems before learning to evaluate, we also use the MetaMATH-Mistral as the base model for MATH-Minos and all other reward models.
For the training of natural language feedback, we use 30k training data generated as Section~\ref{subsec:data_construction} with a learning rate of 5e-6 with a total batch size of 256.
For the training of score feedback, we use 440k training data (i.e., 30k data with the binary classification labels from the training data in the training of natural language feedback and 410k data sampled from MATH-Shepherd.
For the training of baseline, we use the total 440k training data from MATH-Shepherd.
For the training of ORM, we adopt the learning rate of 3e-6 with a batch size of 512. For PRM, the learning rate is 2e-6 with a batch size of 512.
In reinforcement learning, we follow the setting of \citet{wang2024mathshepherd} with the learning rate of 1e-7.

\section{Case Study}
\begin{figure*}[!ht]
    \centering
    \includegraphics[width=1\linewidth]{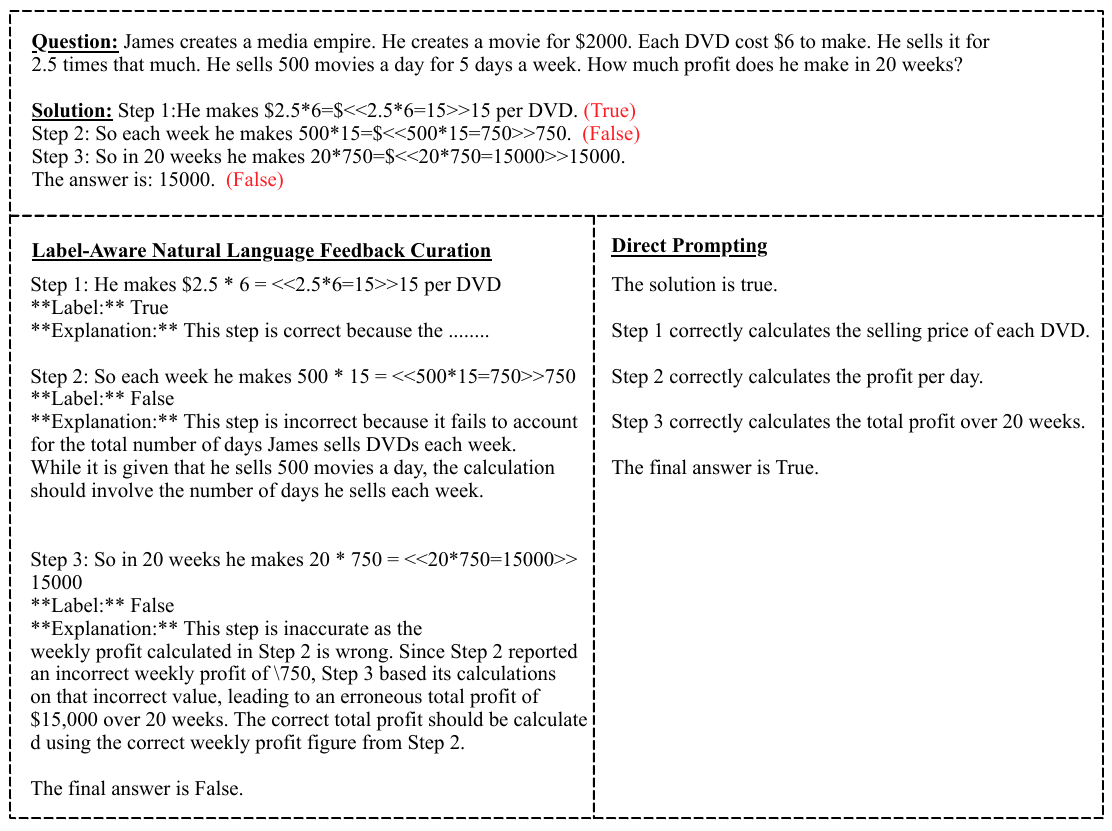}
    \caption{Case study of the Label-Aware Natural Language Feedback Curation.
    }
    \label{fig:case_study}
\end{figure*}

The Figure \ref{fig:case_study} illustrates the different content generated by GPT-4-turbo between Label-Aware Natural Language Feedback Curation and direct prompts. Our proposed method, guided by labels, helps the model mitigate the impact of hallucinations to some extent, resulting in high-quality natural language evaluation.

\clearpage
\begin{figure*}[!ht]
    \centering
    \includegraphics[width=1\linewidth]{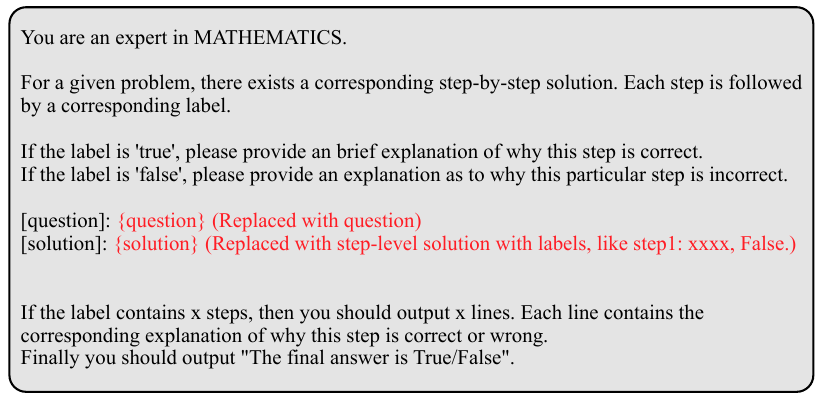}
    \caption{The detailed prompt template used for prompting GPT-4-turbo to generate Label-Aware Natural Language Feedback.
    }
    \label{fig:template_NLF}
\end{figure*}

\begin{figure*}[!ht]
    \centering
    \includegraphics[width=1\linewidth]{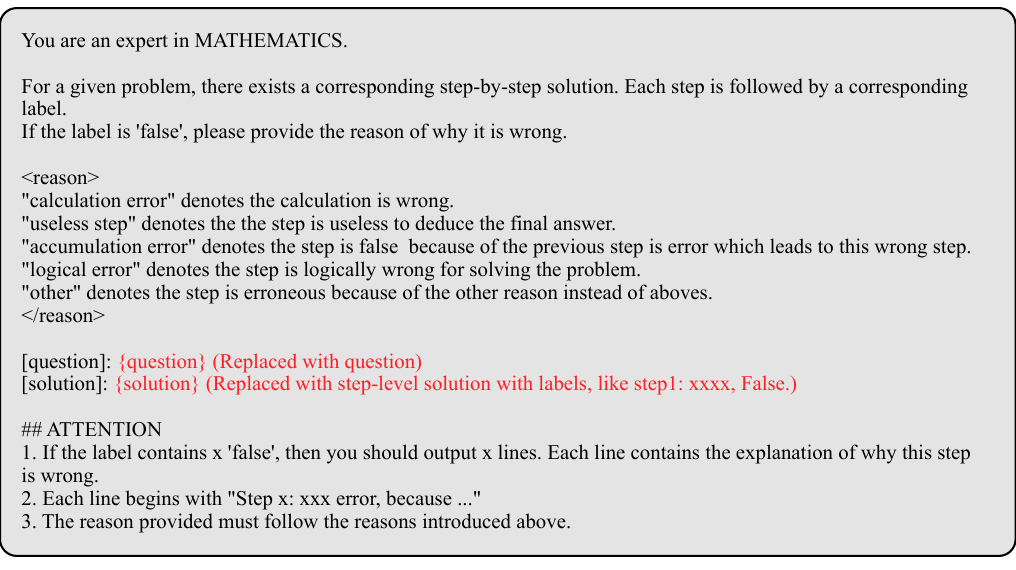}
    \caption{The detailed prompt template used for prompting GPT-4-turbo to generate the error analysis reason.
    }
    \label{fig:template_ANALYSIS}
\end{figure*}

\end{document}